# CausalDisenSeg: A Causality-Guided Disentanglement Framework with Counterfactual Reasoning for Robust Brain Tumor Segmentation Under Missing Modalities

Bo Liu[1,2], Yulong Zou[2], Jin Hong[1,*]

1. School of Information Engineering, Nanchang University, Nanchang 330031, China
2. School of Mathematics and Computer Sciences, Nanchang University, Nanchang 330031, China

E-mail: liuboncu@email.ncu.edu.cn; 9109223090@email.ncu.edu.cn; hongjin@ncu.edu.cn;
* Correspondence should be addressed to Jin Hong

**Abstract**: In clinical practice, the robustness of deep learning models for multimodal brain tumor segmentation is severely compromised by incomplete MRI data. This vulnerability stems primarily from modality bias, where models exploit spurious correlations as shortcuts rather than learning true anatomical structures. Existing feature fusion methods fail to fundamentally eliminate this dependency. To address this, we propose CausalDisenSeg, a novel Structural Causal Model (SCM)-grounded framework that achieves robust segmentation via causality-guided disentanglement and counterfactual reasoning. We reframe the problem as isolating the anatomical Causal Factor from the stylistic Bias Factor. Our framework implements a three-stage causal intervention: (1) Explicit Causal Disentanglement: A Conditional Variational Autoencoder (CVAE) coupled with an HSIC constraint mathematically enforces statistical orthogonality between anatomical and style features. (2) Causal Representation Reinforcement: A Region Causality Module (RCM) explicitly grounds causal features in physical tumor regions. (3) Counterfactual Reasoning: A dual-adversarial strategy actively suppresses the residual Natural Direct Effect (NDE) of the bias, forcing its spatial attention to be mutually exclusive from the causal path. Extensive experiments on the BraTS 2020 dataset demonstrate that CausalDisenSeg significantly outperforms state-of-the-art methods in accuracy and consistency across severe missing-modality scenarios. Furthermore, cross-dataset evaluation on BraTS 2023 under the same protocol yields a state-of-the-art macro-average DSC of 84.49.

## 1. Introduction

In the rapidly evolving landscape of medical image computing, the accurate segmentation of brain tumors, particularly gliomas, stands as a cornerstone of modern neuro-oncology, a field increasingly supported by foundational medical models [1] and robust deep neural networks [2]. Gliomas are among the most aggressive and lethal primary brain malignancies, necessitating precise delineation of tumor sub-regions—including the necrotic core, enhancing tumor, and peritumoral edema—to facilitate effective diagnosis, treatment planning, and prognostic assessment [3]. In clinical practice, multimodal Magnetic Resonance Imaging (MRI) serves as the gold standard for this task, as different pulse sequences provide complementary tissue contrast information: T1-weighted (T1) and T1-contrast enhanced (T1ce) sequences highlight the tumor core structure, while T2-weighted (T2) and Fluid Attenuation Inversion Recovery (FLAIR) sequences are highly sensitive to edema and overall tumor boundaries [4]. Consequently, deep learning approaches—ranging from classic Convolutional Neural Networks (CNNs) such as nnU-Net [5] to emerging Vision Transformers—have achieved remarkable success by fusing these modalities to learn rich, semantically dense representations in full-modality scenarios.

However, the prevailing academic assumption of "data completeness" sharply contrasts with the uncertain reality of clinical workflows. In real-world hospital settings, acquiring a complete set of MRI scans is often precluded by uncontrollable factors such as patient claustrophobia, motion artifacts, inconsistent scanning protocols across institutions, or limited emergency time windows, and financial constraints [6], which has prompted comprehensive literature surveys on this specific challenge [7]. Under such "missing modality" scenarios, state-of-the-art models trained on complete data often suffer catastrophic performance degradation. This fragility stems from the tendency of traditional models to overly rely on the synergy between all modalities during training; once a critical modality (e.g., FLAIR for edema) is absent, the model fails to adapt to the information vacuum, rendering diagnostic predictions unreliable.

To mitigate the impact of missing modalities, existing research has predominantly coalesced around two paradigms: modality synthesis and unified feature learning. Modality synthesis approaches employ Generative Adversarial Networks (GANs) to hallucinate missing sequences from available ones, attempting to complete the input data prior to segmentation [8], alongside recent advancements in multimodal representation learning such as M3AE [9]. While visually appealing, this approach carries high risks in medical contexts, as synthesized images may contain misleading artifacts, and the computational cost of generating 3D volumes prohibits real-time deployment. In contrast, unified feature learning approaches aim to learn a shared latent space robust to input variations. Significant progress has been made, from hetero-modal variational autoencoders [10] and dual disentanglement networks [11] to region-aware fusion networks[12] and shared-specific feature modeling [13], and more recently, methods leveraging prompt learning to interactively extract modality-invariant features [14] or arbitrary cross-modal feature reconstruction [15]. These methods essentially optimize models to maximize the mutual information between available modalities and segmentation targets, utilizing sophisticated attention mechanisms for feature alignment and completion.

Despite these advancements, we argue that existing approaches share a fundamental theoretical blind spot: they rely primarily on statistical correlations rather than causal mechanisms. Neural networks are not just pattern recognizers but are prone to becoming "shortcut learners." From the perspective of causal inference [16, 17], current models easily learn spurious correlations—mapping low-level, modality-specific appearance statistics (e.g., the high-intensity distribution in FLAIR) directly to high-level semantic labels (e.g., tumor), rather than understanding the underlying anatomy. In this context, imaging style (contrast, texture, noise) acts as a Bias Factor, while the true anatomical structure is the Causal Factor. When a specific modality is missing, the model loses this statistical shortcut (the Bias Factor) and fails because it never truly learned to reason based solely on the stable Causal Factor. Current feature fusion methods often aggregate information indiscriminately, whether through simple concatenation, channel exchanging [18], or local expert mixtures [19], exacerbating the entanglement of causal anatomy and spurious style, leaving models vulnerable to distribution shifts.

To transcend the limitations of correlation-based learning, we reframe the robust segmentation task through the lens of Structural Causal Models (SCM) [20] and the foundational principles of causal inference [21]. We conceptualize the image generation and segmentation process as a separated mechanism, drawing inspiration from representation analysis in domain adaptation [22] and invariant conditional distributions [23]: on one hand, the cross-modal invariant anatomical reality, which is the sufficient and necessary cause for the segmentation target; on the other hand, the variable imaging characteristics, which act as confounding noise. A truly robust system must physically isolate these two components and actively block the direct influence of bias on decision-making. Based on this insight, we propose CausalDisenSeg, a causality-guided disentanglement framework designed to learn the interventional distribution via explicit disentanglement and counterfactual intervention. To provide an intuitive overview, Figure 1 illustrates our central hypothesis. Although the four MRI modalities exhibit distinct imaging styles, they share a common anatomical structure that is causally responsible for the segmentation target. Conventional fusion models tend to entangle this stable anatomy with modality-specific appearance bias, thereby learning fragile shortcuts. In contrast, our framework aims to extract the shared causal core across modalities while suppressing modality-dependent bias cues, so that the final prediction is driven by invariant anatomy rather than spurious correlations.

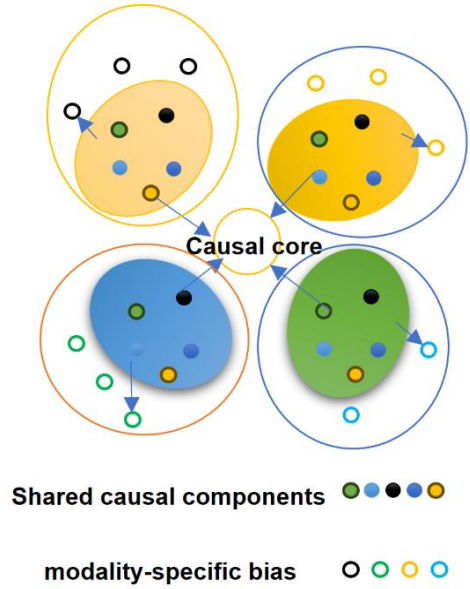

Figure 1 Illustration of the main intuition behind CausalDisenSeg. Different MRI modalities present distinct imaging appearances but preserve a common tumor anatomy. The proposed framework seeks to capture this shared causal anatomy and reduce the influence of modality-specific bias, thereby improving robustness under missing modalities.

Our methodology unfolds in two rigorous stages. First, we implement Explicit Causal Disentanglement. Unlike previous methods that perform simple fusion, we design a Conditional Variational Autoencoder (CVAE) architecture that forces the model to reconstruct original images via separated "anatomical codes" and "style codes." To ensure this separation is not merely superficial, we introduce a fine-grained Hilbert-Schmidt Independence Criterion (HSIC), mathematically enforcing statistical orthogonality between the anatomical and style subspaces. This effectively purifies the feature extraction process from potential bias contamination. Second, even with disentangled features, classifiers may still opportunistically exploit residual bias signals. To address this, we introduce a Counterfactual Reasoning mechanism. We construct a parallel "counterfactual branch" that attempts to predict segmentation results solely from the style bias. We neutralize this branch using a Dual-Adversarial Strategy (Confusion Learning and Discrepancy Learning). This mechanism simulates a counterfactual scenario —"what if only bias existed?"—and actively subtracts this effect during training, forcing the main model to focus exclusively on the anatomical structure anchored by our Region Causality Module. The main contributions of this work are summarized as follows:

(i) Causal Perspective Reframing: We establish a novel structural causal perspective for multimodal

brain tumor segmentation, conceptually distinguishing between the anatomical Causal Factor (target) and the stylistic Bias Factor (confounder), highlighting the theoretical flaws of existing methods in handling spurious correlations.

(ii) HSIC-Constrained Explicit Disentanglement: We propose a rigorous disentanglement module combining CVAE reconstruction with HSIC independence constraints. This ensures not only information completeness but also the strict statistical orthogonality of causal and bias features, cutting off the indirect propagation path of bias at the source.

(iii) Counterfactual Debiasing Mechanism: We design a counterfactual reasoning framework incorporating a Region Causality Module and innovative Confusion and Discrepancy losses. This dual-adversarial mechanism actively estimates and eliminates the direct effect of modality bias during training, ensuring decisions return to the anatomical essence.

(iv) Superior Performance: Extensive experiments on the BraTS 2020 dataset demonstrate that CausalDisenSeg achieves state-of-the-art performance; particularly in extreme missing modality scenarios, our method shows significant advantages in accuracy and consistency compared to recent feature fusion methods. In addition, evaluation on BraTS 2023 under the same protocol confirms stable cross-dataset gains against strong competitor.

## 2. Related work

This section reviews representative methods for incomplete multimodal brain tumor segmentation and causal representation learning, and highlights the key methodological gaps that motivate CausalDisenSeg.

### 2.1. Incomplete Multi-Modal Brain Tumor Segmentation: Paradigms and Limitations

The incomplete availability of multimodal Magnetic Resonance Imaging (MRI) data presents a formidable challenge in clinical neuro-oncology. To address this, existing literature has predominantly bifurcated into Modality Synthesis and Unified Feature Learning. Early generative approaches sought to synthesize missing modalities via Generative Adversarial Networks (GANs) prior to segmentation [8], or through adversarial co-training networks designed to handle missing data [24]. However, these methods incur prohibitive computational costs and risk introducing hallucinatory artifacts that compromise diagnostic reliability. Consequently, the field has pivoted towards Unified Feature Learning, aiming to extract robust latent representations directly from available partial observations.

Foundational works such as HeMIS [25] and U-HVED [10] leveraged hetero-modal variational autoencoders to project diverse modalities into a shared latent space, while others explored style matching strategies to compensate for missing modalities [26]. Building on this, RFNet [12] introduced region-aware fusion strategies to accommodate the heterogeneity of tumor sub-regions. With the advent of Vision Transformers, mmFormer [27] utilized masked modality tokens to model long-range inter-modal dependencies, while M2FTrans [28] proposed a modality-masked fusion transformer to recalibrate features via spatial and channel-wise attention. Similarly, A2FSeg [29] advanced this paradigm by dynamically adapting fusion strategies to varying missing scenarios. In addition, RobustSeg [4] and IM-Fuse [30] further explored robust and inconsistency-aware fusion under heterogeneous subset conditions.

Despite these strides, a fundamental limitation persists: these methods operate under the tacit assumption that augmenting modality inputs invariably yields positive information gain. Challenging this premise, Qiu et al. proposed PNDC [31], demonstrating that uncertain or inconsistent patterns in specific modalities can exert "negative impacts," where direct fusion induces confusion and performance degradation. While PNDC mitigates this via confidence calibration, it—along with A2FSeg and IM-Fuse [30]—remains bound to feature-level weighting or post-hoc correction, bypassing formal predictive uncertainty estimations like Bayesian approximation or deep ensembles [32, 33]. Crucially, these methods fail to sever the model's reliance on spurious correlations between modality-specific attributes (e.g., intensity distributions) and semantic labels. Unlike PNDC and A2FSeg, which focus on feature interaction, our approach seeks to eliminate these dependencies at the source through a structural causal mechanism.

**2.2. Causal Representation Disentanglement and Counterfactual Reasoning**

Causal Inference has emerged as a transformative paradigm in computer vision, transcending mere statistical association to uncover the underlying mechanisms of data generation. In the context of Domain Generalization (DG), the objective is often framed as disentangling Causal Factors (invariant semantics) from Non-causal Factors (domain-specific styles or bias).

To achieve this separation, independence constraints have become a standard methodology. For instance, Chen et al. [16] proposed a causal-inspired early-branching architecture, leveraging the Hilbert-Schmidt Independence Criterion (HSIC) [34] to mathematically enforce statistical orthogonality between semantic and domain-specific feature subspaces. Furthermore, Counterfactual Reasoning serves as a powerful tool for debiasing by estimating and removing the Natural Direct Effect (NDE) of confounders.

While causal frameworks have proven effective in sentiment analysis and general visual recognition, to the best of our knowledge, no prior work has systematically applied causal inference to the specific problem of incomplete multi-modal brain tumor segmentation. Existing segmentation models remain susceptible to "statistical shortcuts" formed by Bias Factor.

CausalDisenSeg bridges this gap. Diverging from methods that manipulate entangled features, we establish the first Structural Causal Model (SCM) tailored for missing modalities in brain tumor segmentation. We introduce a rigorous disentanglement mechanism combining Conditional VAEs with HSIC constraints to enforce the separation of anatomical structure from imaging style. Furthermore, we innovate a spatial counterfactual reasoning mechanism that actively suppresses the direct influence of modality bias. By physically severing the shortcut pathways, our framework ensures robust segmentation driven solely by stable anatomical causal factors, even in scenarios with severe modality dropout.

# 3. Method

In this section, we present CausalDisenSeg, a comprehensive framework designed to ensure robust brain tumor segmentation under missing modalities. We first rigorously formulate the problem through the lens of a Structural Causal Model (SCM), identifying the specific causal pathways that introduce modality bias. Subsequently, we detail the implementation of our three core components as illustrated in Figure 3: (1) Causality-Guided Disentanglement via Conditional Variational Autoencoders (CVAE) and Hilbert-Schmidt Independence Criterion (HSIC); (2) Causal Representation Reinforcement via the Region Causality Module (RCM); and (3) Counterfactual Reasoning for eliminating the Natural Direct Effect (NDE) of bias through dual-adversarial learning.

**3.1. Problem Formulation and Structural Causal Model**

We begin by formalizing incomplete-modality segmentation under a structural causal framework, which makes the bias pathways explicit and provides a principled target for intervention during training.

*3.1.1. Notation and Problem Definition*

Let $\mathcal{D} = \{(X_i, Y_i)\}_{i=1}^{N}$ denote a multimodal MRI dataset, where $X_i = \{x_i^m\}_{m \in M}$ represents the set of available modality volumes (e.g., T1, T1ce, T2, FLAIR) for the i-th patient, and $Y_i \in \{0,1\}^{H \times W \times D}$ represents the corresponding pixel-wise ground truth segmentation mask. In clinical practice, the full set of modalities $M_{full}$ is often unavailable; instead, we observe a subset $M_{obs} \subset M_{full}$.

The objective of traditional deep learning models is to learn a mapping function $f: X \to Y$ that minimizes the empirical risk. However, standard correlation-based learning inevitably captures spurious correlations between modality-specific appearances (e.g., intensity distributions) and labels. To address this, we reframe the task as a causal inference problem.

*3.1.2. Structural Causal Model (SCM)*

We propose a Structural Causal Model (SCM) to describe the data generation and inference mechanism for multimodal segmentation. As illustrated in Figure 2 (Causal Graph), we decompose the observed variables into latent factors governed by the following causal relationships:

(i) Causal Factor ($C$): This latent variable represents the intrinsic anatomical structure of the brain tumor (e.g., the spatial location and shape of the necrosis or edema). $C$ is modality-invariant and serves as the sufficient and necessary cause for the segmentation label $Y$.

(ii) Bias Factor ($B$): This latent variable encapsulates the modality-specific imaging characteristics, such as contrast levels, noise patterns, and texture styles unique to each MRI sequence (e.g., the high signal intensity of CSF in T2). $B$ acts as a confounder in the system.

(iv) Fused Multimodal Representation ($M$): We define $M$ as the Mediator. In a deep neural network, this corresponds to the high-level feature representation obtained after fusing individual modality encodings. $M$ is the direct precursor to the final prediction.

(v) Prediction ($Y$): The final segmentation output generated by the model.

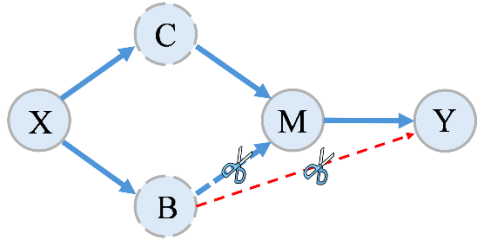

Figure 2 The proposed Structural Causal Model (SCM) for robust brain tumor segmentation. We identify two confounding pathways that compromise robustness: the Indirect Bias Path (X → B → M) and the Natural Direct Effect (B → Y). Our framework applies a dual-intervention strategy (indicated by scissors) to structurally sever the dependency of representations on modality bias and functionally eliminate shortcut learning.

*3.1.3. Decomposition of Causal Pathways*

Based on the SCM in Figure 2, the information flow from the input $X$ to the prediction $Y$ propagates through three distinct pathways. Understanding these pathways is crucial for designing our debiasing strategy:

(i) Path 1: The Causal Path ( $X \to C \to M \to Y$)

This is the ideal decision pathway. The model correctly extracts the causal structure ($C$), integrates it into a robust representation ($M$), and generates the prediction ($Y$). In causal terms, this path corresponds to the Total Indirect Effect (TIE) driven by anatomy. Our goal is to maximize the reliance on this path.

(ii) Path 2: The Indirect Bias Path ($X \to B \to M \to Y$)

This path represents feature contamination. In standard encoders, the Bias Factor ($B$) is not explicitly separated from $C$. Consequently, modality-specific styles leak into the fused representation ($M$). For instance, the model might encode "high brightness" (a feature of $B$) into $M$ rather than "edema structure" (a feature of $C$). This makes $M$ unstable: if the specific modality providing that brightness is missing, $M$ collapses.

(iii) Path 3: The Direct Shortcut Path ($X \to B \to Y$)

This path represents shortcut learning and corresponds to the Natural Direct Effect (NDE) of the bias. Here, the model bypasses the complex representation learning of $M$ and directly exploits low-level statistical irregularities in $B$ to overfit the ground truth. For example, the model might learn a heuristic rule: "if T1ce intensity > threshold, then tumor core," without understanding the shape. This path is highly fragile to domain shifts and missing modalities.

*3.1.4. The Intervention Objective*

Our theoretical objective is to learn a prediction model that approximates the interventional distribution $P(Y|do(C))$. The do-operator signifies a physical intervention where we force the Causal Factor to take a value while severing its dependence on confounders.

Mathematically, the Total Effect (TE) of the input on the prediction can be decomposed as:

$$TE = TIE(C \to M \to Y \text{ and } B \to M \to Y) + NDE(B \to Y) \tag{1}$$

To achieve robust segmentation, we must eliminate the NDE (the direct shortcut) and block the Indirect Bias Path (B $\to$ M), ensuring that the final decision is purely driven by the anatomical structure:

$$Y_{pred} \approx f(C) \quad s.t. \quad C \perp\!\!\!\perp B \tag{2}$$

**3.2. Explicit Feature Disentanglement via Conditional VAE**

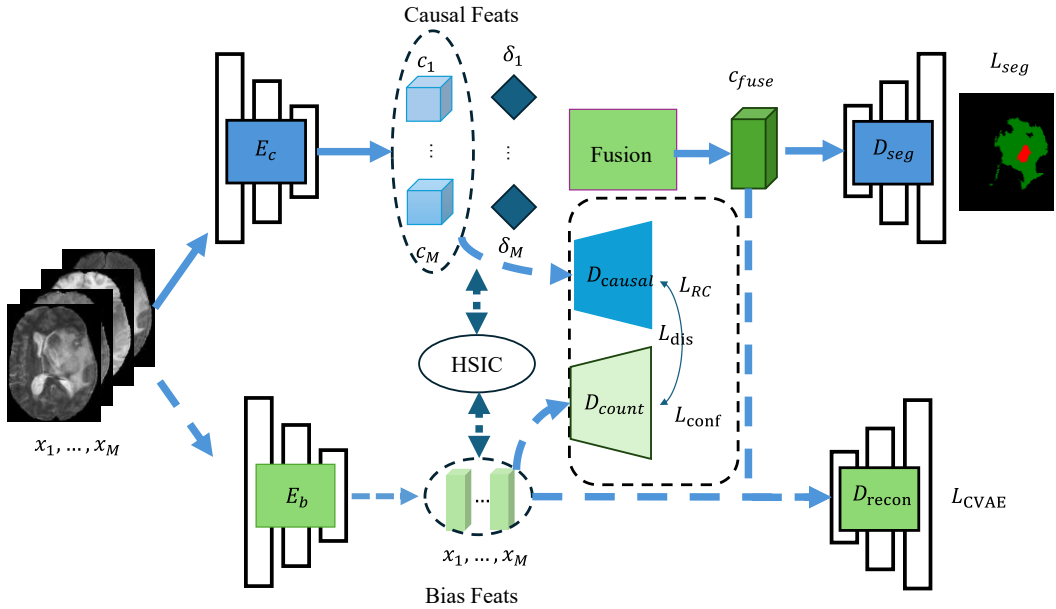

Figure 3 The overall architecture of CausalDisenSeg. The framework is organized into three functional branches: (1) The Causal Stream (Top) extracts anatomical features ($c_M$) via $E_c$ to generate the segmentation mask ($D_{seg}$), reinforced by the Region Causality Module (RCM) to anchor attention on tumor regions. (2) The Bias Stream (Bottom) captures modality styles ($b_M$) via

$E_b$ and, together with $c_M$, drives the Conditional Reconstruction Decoder ($D_{recon}$) to ensure information completeness. (3) The Counterfactual Branch (Middle) employs a dual-adversarial strategy ($D_{count}$) to neutralize the Natural Direct Effect (NDE) of bias. Note: During inference, only the clean Causal Stream (Top) is retained.

As visualized in the overall architecture (Figure 3), our framework processes the input modalities $(x_1, ..., x_M)$ through two distinct parallel streams: the Causal Stream (Top) and the Bias Stream (Bottom). To physically isolate the Causal Factor ($C$) from the Bias Factor ($B$) as defined in our SCM, we propose a Causality-Guided Disentanglement Module. Unlike previous methods that rely on implicit feature separation, we employ a Conditional Variational Autoencoder (CVAE) architecture to enforce semantic distinctness through reconstruction.

*3.2.1. Dual-Encoder Architecture*

For an input volume $x_m$ from modality m, we utilize two distinct encoders:

(i) Causal Encoder ($E_c$): Located in the Causal Stream (Figure 3, Top, Blue Block), this U-Net style encoder, built upon the foundational 3D U-Net [35] volumetric architecture, extracts spatial-structural features. We denote the output feature map as the Causal Factor $c_m \in R^{C \times D \times H \times W}$.

$$c_m = E_c(x_m) \tag{3}$$

(ii) Bias Encoder ($E_{bias}$): Located in the Bias Stream (Figure 3, Bottom, Green Block), this lightweight convolutional network designed to capture global style information. Crucially, to prevent this encoder from retaining spatial structure (which belongs to $C$), we apply Global Average Pooling (GAP) at the final layer. The output is the bias feats $b_m \in R^L$, a low-dimensional latent vector (represented as green vectors in Figure 3).

$$b_m = GAP(E_{bias}(x_m)) \tag{4}$$

*3.2.2. Conditional Reconstruction Mechanism*

As shown in the center of Figure 3, the extracted Causal Feats ($c_1 ... c_M$) are aggregated alongside modality indicators $\delta_1 ... \delta_M$ (depicted as diamonds) into the Fusion Module. This produces a unified representation $c_{fuse}$.

To enforce disentanglement, we introduce a Conditional Reconstruction Decoder ($D_{recon}$), shown on the bottom right of Figure 3. The goal of $D_{recon}$ is to reconstruct the original image $x_m$ by combining the robust $c_{fuse}$ and the modality-specific "bias" from $b_m$.

The reconstructed volume $\widehat{x_m}$ is generated by the decoder via the dashed pathways connecting Bias Feats and $c_{fuse}$ to $D_{recon}$:

$$\widehat{x_m} = D_{recon}\left(AdaIN(c_{fuse}, b_m)\right) \tag{5}$$

*3.2.3. CVAE Optimization Objective*

We formulate the learning objective as maximizing the Evidence Lower Bound (ELBO) of the data likelihood. The loss function for the disentanglement module, $L_{CVAE}$, consists of a reconstruction term and a regularization term:

$$L_{CVAE} = \sum_{m \in M_{obs}} \left( \underbrace{|x_m - \widehat{x_m}|_1}_{Reconstruction\ Loss} + \lambda_{KL} \underbrace{D_{KL}\left(q_\phi(b_m|x_m)|p(b)\right)}_{KL\ Divergence} \right) \tag{6}$$

(i) Reconstruction Loss: Ensures that $c_m$ (causal) and $b_m$ (bias) together preserve all necessary

information to reproduce the input. If $E_{main}$ ignores causal or $E_{bias}$ ignores bias, reconstruction fails.

(ii) KL Divergence: Constrains the posterior distribution of the Bias Factor $q_\phi(b_m|x_m)$ to approximate a standard normal prior $p(b) \sim \mathcal{N}(0, I)$. This regularization ensures the bias latent space is smooth and compact.

### 3.3. Statistical Independence via HSIC (Cutting the Indirect Bias Path)

The CVAE ensures information completeness, but it does not guarantee exclusivity. There is a risk that the Main Encoder $E_{main}$ leaks style information into $c_m$, or $E_{bias}$ captures residual structure. If this happens, the Indirect Bias Path ($X \to B \to M \to Y$) remains active, as bias flows into the Mediator $M$.

To explicitly sever this link ($B \to M$), we impose a statistical independence constraint using the Hilbert-Schmidt Independence Criterion (HSIC). As indicated by the "HSIC" oval connecting the Causal Feats and Bias Feats in Figure 3, HSIC is a robust, kernel-based measure of dependence that assesses both linear and non-linear correlations between random variables.

*3.3.1. Kernel-Based Independence*

We map the Causal Factors $C = \{c_m\}$ and Bias Factors $B = \{b_m\}$ into Reproducing Kernel Hilbert Spaces (RKHS), denoted as $H_C$ and $H_C$. The HSIC value is the squared Hilbert-Schmidt norm of the cross-covariance operator $\Sigma_{CB}$:

$$HSIC(P_{CB}, H_C, H_B) = |\Sigma_{CB}|^2_{HS} \tag{7}$$

If $HSIC(C, B) = 0$, then $C$ and $B$ are statistically independent ($C \perp\!\!\!\perp B$).

*3.3.2. Empirical HSIC Loss*

Since linear independence does not imply statistical independence for non-Gaussian variables, we map features to a Reproducing Kernel Hilbert Space (RKHS). We employ the Gaussian Radial Basis Function (RBF) kernel $k(u, v) = exp(-|u - v|^2/\sigma^2)$ to construct the Gram matrices $K_C$ and $K_B$. Minimizing the Hilbert-Schmidt norm of the cross-covariance operator in this RKHS is equivalent to minimizing the mutual information between $C$ and B.

For a mini-batch of size $N$, let $K_C \in R^{N \times N}$ and $K_B \in R^{N \times N}$ be the Gram matrices computed using a Radial Basis Function (RBF) kernel, where $(K_C)_{ij} = k(c_i, c_j) and (K_B)_{ij} = k(b_i, b_j)$. The empirical HSIC loss is defined as:

$$\mathcal{L}_{HSIC} = \frac{1}{(N-1)^2} Tr(K_C H K_B H) \tag{8}$$

where $H = I - \frac{1}{N}$ is the centering matrix. By minimizing $\mathcal{L}_{HSIC}$, we mathematically enforce orthogonality between the causal and bias subspaces. This ensures that the Causal Factor C (and consequently the Mediator M) contains zero mutual information with the Bias Factor B, effectively purifying the feature representation used for downstream segmentation.

### 3.4. Causal Representation Reinforcement (Region Causality Module)

Severing bias paths is insufficient if the "Causal Factor" $C$ does not actually encode relevant anatomy. To anchor the causal path ($X \to C \to M \to Y$) to physical reality, we introduce the Region Causality Module, implemented as a decoder branch $D_{causal}$ in Figure 3.

As shown in Figure 3 (Framework), the fused multimodal representation $M$ (output of the

Transformer bottleneck) serves as the input to the RCM. The RCM acts as a "causal probe," projecting the high-dimensional features into a single-channel spatial attention map, termed the Causality Map ($A_{causal}$):

$$A_{causal} = \sigma(Conv_{1\times1}(M)) \qquad (9)$$

where $\sigma$ is the sigmoid function. $A_{causal} \in [0,1]^{H \times W \times D}$ represents the model's spatial attention based on the causal features.

We explicitly supervise this map using the ground truth mask $Y_{GT}$ to ensure it highlights the tumor regions:

$$\mathcal{L}_{\mathcal{RC}} = \mathcal{L}_{\mathcal{D}ice}(A_{causal}, Y_{GT}) + \mathcal{L}_{\mathcal{BCE}}(A_{causal}, Y_{GT}) \qquad (10)$$

By minimizing $\mathcal{L}_{\mathcal{RC}}$, we force the Mediator $M$ to encode highly localized, structure-aware information. Crucially, this map $A_{causal}$ serves as a spatial reference for the subsequent counterfactual discrepancy learning.

**3.5. Counterfactual Reasoning: Eliminating the Natural Direct Effect**

Even with perfect disentanglement and reinforcement, a deep classifier might still discover residual statistical shortcuts in the bias features, keeping the Direct Shortcut Path ($X \rightarrow B \rightarrow Y$) active. To eliminate this Natural Direct Effect (NDE), we employ a Counterfactual Reasoning framework.

*3.5.1. The Counterfactual Scenario*

We simulate a counterfactual world: "What would the model predict if it had access only to the Bias Factor and no causal knowledge?"

To model this, we construct a Counterfactual Branch with a lightweight predictor $D_{count}$. This branch receives only the Bias Factors $B_{all} = \{b_1, \ldots, b_m\}$ (concatenated from all available modalities) as input and outputs a "bias-only" prediction $Y_{bias}$:

$$Y_{bias} = D_{count}(B_{all}) \qquad (11)$$

If the NDE is non-zero, $Y_{bias}$ will show some correlation with the ground truth. Our goal is to suppress this capability using a Dual-Adversarial Strategy: Confusion Learning and Mutually Exclusive Learning.

*3.5.2. Confusion Learning (Uniform Constraint)*

Since the Bias Factor $B$ is defined as global and non-structural, it should theoretically contain no information about the specific location of the tumor. Therefore, any predictive power in $Y_{bias}$ implies leakage.

We enforce the removal of this information by maximizing the entropy of the counterfactual prediction. We define a Confusion Loss ($\mathcal{L}_{conf}$) that forces $Y_{bias}$ to approximate a Uniform Distribution ($\mathcal{U}$):

$$\mathcal{L}_{conf} = MSE(Softmax(Y_{bias}), \mathcal{U}) \qquad (12)$$

where $\mathcal{U}$ is a tensor filled with $1/K$ ($K$ being the number of classes). Minimizing this loss penalizes the Bias Encoder if $B$ retains any discriminative cues, effectively "confusing" the bias branch. This drives the NDE towards zero by stripping the bias factors of their predictive utility.

*3.5.3. Mutually Exclusive Learning (Discrepancy Constraint)*

Confusion learning handles global statistics, but we must also ensure that the bias branch does not focus on the same spatial regions as the causal branch.

We leverage the Causality Map ($A_{causal}$) generated by the RCM (Section 3.4) and the activation map of the bias branch $\widehat{Y_{bias}}$. We define a Discrepancy Loss ($\mathcal{L}_{dis}$) using the Cosine Similarity metric:

$$\mathcal{L}_{dis} = CosineSim\left(vec(A_{causal}), vec(\widehat{Y_{bias}})\right) = \frac{\langle A_{causal}, \widehat{Y_{bias}} \rangle}{|A_{causal}|_2 \cdot |\widehat{Y_{bias}}|_2} \quad (13)$$

By minimizing this similarity, we enforce a spatial orthogonality constraint: the regions utilized by the bias path must be disjoint from the true tumor regions identified by the causal path.

(i) Since $A_{causal}$ is supervised to overlap with the tumor ($\mathcal{L}_{RC}$), minimizing the overlap between $A_{causal}$ and $\widehat{Y_{bias}}$ effectively pushes the bias attention away from the tumor.

(ii) This physical separation ensures that the final segmentation Y cannot be derived from the bias features, thereby physically cutting the Direct Shortcut Path ($B \rightarrow Y$).

While $L_{conf}$ handles the statistical distribution output, $L_{dis}$ enforces a spatial constraint. It forces the bias branch to attend to background or noise regions, physically preventing it from overlapping with the tumor regions identified by the RCM.

### 3.6. Total Optimization Objective and Inference

After defining each component loss, we summarize the joint objective used for end-to-end training and clarify the lightweight inference path retained at test time.

*3.6.1. Joint Training*

The entire CausalDisenSeg framework is trained end-to-end. The total objective function $\mathcal{L}_{total}$ is a weighted sum of the segmentation loss and the causal intervention losses:

$$\mathcal{L}_{total} = \mathcal{L}_{seg}(Y_{pred}, Y_{GT}) + \lambda_1 \mathcal{L}_{CVAE} + \lambda_2 \mathcal{L}_{HSIC} + \lambda_3 \mathcal{L}_{RC} + \lambda_4 \mathcal{L}_{conf} + \lambda_5 \mathcal{L}_{dis} \quad (14)$$

where:

(i) $\mathcal{L}_{seg}$ is the standard combination of Dice Loss and Cross-Entropy Loss applied to the main output.

(ii) $\lambda_1, \ldots, \lambda_5$ are hyperparameters balancing the contributions of disentanglement, independence, reinforcement, and counterfactual reasoning.

*3.6.2. Causal Inference*

During the inference phase, the auxiliary components designed for training intervention—specifically the Bias Encoders ($E_{bias}$), the VAE Decoders ($D_{rec}$), and the Counterfactual Branch ($P_{bias}$)—are discarded. This asymmetrical architecture allows us to utilize the bias stream for 'model debugging' during training, while maintaining a pure, lightweight causal stream for inference.
The model operates solely using the robust Causal Path:

$$X \xrightarrow{E_c} C \xrightarrow{Fusion} M \xrightarrow{D_{seg}} Y_{pred} \quad (15)$$

Since the Main Encoder ($E_c$) and the Mediator ($M$) have been purified of bias via $\mathcal{L}_{HSIC}$ and $\mathcal{L}_{dis}$ during training, the inference process relies exclusively on the stable anatomical representations. This allows the model to maintain high performance even when specific modalities (and their associated biases) are missing, as the decision logic is grounded in the invariant causal structure rather than transient imaging styles.

## 4. Experiments

We conduct comprehensive experiments to evaluate segmentation accuracy, robustness under heterogeneous missing-modality patterns, and the contribution of each causal component in the proposed framework.

### 4.1. Dataset and Evaluation Metrics

To comprehensively evaluate the effectiveness of CausalDisenSeg, we conduct extensive experiments on the BraTS 2020 and BraTS 2023 datasets, which serve as widely recognized benchmarks for multimodal brain tumor segmentation [36]. Each subject within these datasets comprises multiparametric MRI scans featuring four distinct structural modalities: T1-weighted (T1), contrast-enhanced T1-weighted (T1ce/T1Gd), T2-weighted (T2), and Fluid Attenuation Inversion Recovery (FLAIR/T2-FLAIR). For the BraTS 2020 dataset, which contains 369 training cases, we partition the data into training, validation, and testing sets following the established schemes detailed in RFNet [12]. To assess cross-dataset robustness, the BraTS 2023 dataset, which contains 1251 cases, is randomly partitioned into 875 training, 125 validation, and 251 testing cases, strictly following a 7:1:2 ratio.

To rigorously evaluate performance under missing modality scenarios, we adopt the standard protocol established in RFNet [12], ensuring consistent and fair comparisons with existing literature. The quantitative evaluation of our segmentation performance is conducted using solely the Dice Similarity Coefficient (DSC). Specifically, we report the DSC across three clinical target regions: the Whole Tumor (WT), the Tumor Core (TC), and the Enhancing Tumor (ET). These regions are delineated by amalgamating specific tumor sub-structures—namely the necrotic and non-enhancing tumor core (NCR/NET), peritumoral edema (ED), and GD-enhancing tumor (ET)—with the background (BG). A higher DSC value indicates a greater spatial overlap between the model's prediction and the expert-annotated ground truth, directly reflecting superior segmentation accuracy.

### 4.2. Implementation Details

The proposed framework is implemented using PyTorch and trained on an NVIDIA RTX 4090 GPU. We employ the Adam optimizer with an initial learning rate of $2\times10^{-4}$ and a weight decay of $1\times10^{-5}$. The learning rate is scheduled using a cosine annealing strategy. The trade-off hyperparameters in the loss function are empirically set to $\lambda_1=0.1$, $\lambda_2=0.1$, $\lambda_3=1.0$, $\lambda_4=0.5$, and $\lambda_5=0.5$. To simulate missing modalities during the training phase, we employ a Bernoulli sampling strategy where each modality is dropped with a probability of 0.5, ensuring the model learns to adapt to incomplete inputs.

### 4.3. Comparison with State-of-the-Art Methods

We compare CausalDisenSeg against a comprehensive set of state-of-the-art methods, ranging from foundational generative approaches to the latest feature fusion and calibration frameworks. The baselines include: Representation Learning Methods: HeMIS [25] and U-HVED [10]. Feature Fusion Networks: RobustSeg [4], RFNet [12], mmFormer [27], and IM-Fuse [30]. Recent Advanced Frameworks: A2FSeg [29] (Adaptive fusion), and PNDC [31] (Positive-Negative impact calibration).

As summarized in Table 1, CausalDisenSeg achieves the best average DSC across all three tumor regions: 87.84 (WT), 79.54 (TC), and 62.75 (ET). Compared with the strongest non-causal competitors under the same protocol, the gains are +0.61 DSC on WT (vs. PNDC [31]), +0.53 DSC on TC (vs. IM-Fuse [30]), and +0.61 DSC on ET (vs. PNDC [31]).

To evaluate overall robustness, we further average the three regional averages, obtaining 76.71 for CausalDisenSeg, compared with 76.02 for PNDC [31] and 75.93 for IM-Fuse [30]. While absolute dominance across every isolated modality subset is theoretically constrained by the inherent information bottleneck, CausalDisenSeg demonstrates unparalleled stability and aggregate performance under heterogeneous missing-modality patterns. This empirical evidence strongly validates our causal hypothesis: mitigating shortcut learning yields more resilient representations compared to merely overfitting to specific modality combinations.

Table 1 Performance comparison (DSC%) with SOTA methods on BraTS 2020. Available and missing modalities are denoted by ✓ and ✗, respectively.

| M | | | | | | | | | | | | | | | | | |
|---|---|---|---|---|---|---|---|---|---|---|---|---|---|---|---|---|---|
| | FLAIR | ✓ | ✗ | ✗ | ✗ | ✓ | ✓ | ✓ | ✗ | ✗ | ✗ | ✓ | ✓ | ✓ | ✗ | ✓ | |
| | T1ce | ✗ | ✓ | ✗ | ✗ | ✓ | ✗ | ✗ | ✓ | ✓ | ✗ | ✓ | ✓ | ✗ | ✓ | ✓ | AVG |
| | T1 | ✗ | ✗ | ✓ | ✗ | ✗ | ✓ | ✗ | ✓ | ✗ | ✓ | ✓ | ✗ | ✓ | ✓ | ✓ | |
| | T2 | ✗ | ✗ | ✗ | ✓ | ✗ | ✗ | ✓ | ✗ | ✓ | ✓ | ✗ | ✓ | ✓ | ✓ | ✓ | |
| WT | HeMIS [25] | 71.6 | 67.71 | 68.96 | 68.19 | 69.17 | 68.67 | 69.83 | 69.01 | 69.78 | 69.4 | 70.21 | 71.28 | 70.73 | 71.58 | 72.06 | 69.88 |
| | U-HVED [10] | 69.85 | 46.82 | 46.77 | 54.03 | 61.45 | 58.25 | 64.5 | 62.91 | 65.76 | 64.29 | 66.99 | 69.7 | 68.38 | 70.35 | 71.41 | 62.76 |
| | RFNet [12] | 86.42 | 77.34 | 76.46 | 86.21 | 89.55 | **89.3** | 89.35 | 81 | 87.45 | 87.95 | 90.39 | 90.20 | 90.42 | 88.59 | 90.77 | 86.76 |
| | mmFormer [27] | 82.4 | 74.25 | 74.37 | 83.07 | 84.54 | 84.61 | 85.82 | 77.98 | 84.05 | 84 | 85.34 | 86.11 | 86.22 | 84.64 | 86.38 | 82.92 |
| | A2FSeg [29] | 85.61 | 75.36 | 75.35 | 86.45 | 87.53 | 79.97 | 89.16 | 87.35 | 89.23 | 89.21 | 89.78 | 90.05 | 89.98 | 88.06 | 90.45 | 86.24 |
| | IM-Fuse [30] | **87.55** | 76.9 | 77.52 | 86.29 | **89.57** | 89.81 | 89.9 | 81.28 | 88.01 | 88.03 | 90.44 | 90.44 | 90.73 | 88.79 | 91.02 | 87.08 |
| | PNDC [31] | 86.66 | 77.5 | 77.24 | 87.24 | 88.29 | 81.54 | 90.21 | 88.02 | 90.04 | 89.99 | 90.52 | 90.44 | 90.65 | 88.95 | 91.07 | 87.23 |
| | **Ours** | 87.31 | **78.83** | **78.42** | **87.72** | 88.77 | 82.51 | **90.87** | **88.44** | **90.54** | **90.48** | **90.98** | **90.68** | **91.07** | **89.51** | **91.46** | **87.84** |
| TC | HeMIS [25] | 53.43 | 51.41 | 51.56 | 51.11 | 51.7 | 51.08 | 51.85 | 51.88 | 52.35 | 51.51 | 52.95 | 53.76 | 52.97 | 54.38 | 55.03 | 52.46 |
| | U-HVED [10] | 34.62 | 35.51 | 27.3 | 37.67 | 42.15 | 38.26 | 43.41 | 44.93 | 47.53 | 44.97 | 49.13 | 51.30 | 49.40 | 52.72 | 54.17 | 43.53 |
| | RFNet [12] | 65.04 | 82.37 | 64.31 | 68.47 | 84.69 | 71.45 | 72.62 | 83.15 | 84.06 | 72.11 | 84.71 | **84.70** | 74.28 | 84.11 | 84.74 | 77.39 |
| | mmFormer [27] | 66.19 | 77.96 | 61.17 | 69.18 | 80.36 | 69.58 | 71.55 | 79.93 | 80.79 | 70.90 | 80.18 | 81.31 | 72.02 | 81.12 | 81.22 | 74.9 |
| | A2FSeg [29] | 70.91 | 81.65 | 65.4 | 69.53 | 84.63 | 82.80 | 72.89 | 72.73 | 73.70 | 84.93 | 85.13 | 74.61 | 85.00 | 84.65 | 84.95 | 78.23 |
| | IM-Fuse [30] | 70.18 | 83.09 | 66.46 | 71.22 | 85.15 | 73.90 | 73.81 | **84.64** | **85.46** | 73.88 | 85.5 | 85.21 | 75.50 | 85.74 | 85.51 | 79.01 |
| | PNDC [31] | 71.36 | 82.23 | 66.13 | 70.63 | 84.84 | 83.31 | 73.57 | 72.82 | 74.26 | 85.24 | 85.47 | 75.06 | 85.24 | 84.89 | 85.24 | 78.68 |
| | **Ours** | **72.21** | **83.34** | **67.53** | **72.73** | **85.24** | **84.28** | **74.85** | 72.96 | 75.32 | **85.84** | **86.12** | 75.91 | **85.69** | **85.34** | **85.79** | **79.54** |
| ET | HeMIS [25] | 43.77 | 42.41 | **41.59** | 41.45 | 41.83 | 40.29 | 41.19 | 42.08 | 42.39 | 41.00 | 43.67 | 44.16 | 42.95 | 45.27 | 46.33 | 42.69 |
| | U-HVED [10] | 12.88 | 24.94 | 7.27 | 24.26 | 30.02 | 21.95 | 29.4 | 33.64 | 36.18 | 32.12 | 39.39 | 40.91 | 38.09 | 43.18 | 45.33 | 30.64 |
| | RFNet [12] | 40.47 | 74.27 | 37.51 | 43.59 | 76.45 | 43.81 | 46.99 | 75.22 | **73.94** | 46.37 | 77.01 | **76.38** | 48.95 | 76.38 | 76.64 | 60.93 |
| | mmFormer [27] | 40.47 | 68.91 | 33.97 | 45.61 | 69.81 | 43.63 | 48.09 | 71.10 | 70.72 | 45.92 | 70.08 | 71.6 | 48.38 | 70.65 | 71.36 | 58.02 |
| | A2FSeg [29] | 47.27 | 76.54 | 35.47 | 39.95 | 75.73 | 77.55 | 42.47 | 48.34 | 48.91 | 77.26 | 77.34 | 48.96 | **76.11** | 76.54 | 78.17 | 61.77 |
| | IM-Fuse [30] | 41.21 | 73.11 | 36.25 | **47.37** | 74.81 | 45.41 | **49.09** | **76.48** | 76.06 | 49.24 | 76.68 | 76.29 | 50.03 | **77.00** | 76.63 | 61.70 |
| | PNDC [31] | 48.19 | 76.62 | 36.02 | 41.18 | 76.08 | 77.89 | 43.51 | 48.55 | 49.65 | 77.30 | 77.50 | 49.12 | 75.90 | 76.42 | 78.23 | 62.14 |
| | **Ours** | **49.71** | **76.74** | 36.93 | 43.25 | **76.65** | **78.46** | 45.23 | 48.90 | 50.87 | **77.35** | **77.78** | 49.37 | 75.56 | 76.21 | **78.31** | **62.75** |

### 4.3.1. Performance on BraTS 2023

To evaluate cross-dataset robustness, we additionally benchmark all methods on BraTS 2023 under the same 15 missing-modality settings. As shown in Table 2, CausalDisenSeg achieves the best average DSC in all three tumor regions: 89.99 (WT), 86.89 (TC), and 76.59 (ET). Compared with the strongest baseline ([31] for WT/TC/ET), the gains are +0.25, +1.26, and +0.49 DSC, respectively. The macro-average over WT/TC/ET improves from 83.82 to 84.49.

A finer-grained subset analysis shows that our method is strongest on TC and remains competitive on WT and ET. This pattern indicates that causal constraints provide stable gains on structurally informative regions while some highly sparse modality combinations remain challenging, especially for ET where enhancement cues are the most vulnerable to modality incompleteness.

Table 2 Performance comparison (DSC%) on BraTS 2023 across 15 missing-modality settings.

| M | | | | | | | | | | | | | | | | | | AVG |
|---|---|---|---|---|---|---|---|---|---|---|---|---|---|---|---|---|---|---|
| | FLAIR | ✓ | × | × | × | ✓ | ✓ | ✓ | × | × | × | ✓ | ✓ | ✓ | × | ✓ | | |
| | T1ce | × | ✓ | × | × | ✓ | × | × | ✓ | ✓ | × | ✓ | ✓ | × | ✓ | ✓ | | |
| | T1 | × | × | ✓ | × | × | ✓ | × | ✓ | × | ✓ | ✓ | × | ✓ | ✓ | ✓ | | |
| | T2 | × | × | × | ✓ | × | × | ✓ | × | ✓ | ✓ | × | ✓ | ✓ | ✓ | ✓ | | |
| WT | U-HVED [10] | 83.12 | 77.28 | 73.52 | 80.18 | 81.64 | 82.98 | 79.63 | 70.35 | 82.68 | 78.45 | 82.12 | 83.65 | 83.16 | 78.23 | 84.45 | 80.10 |
| | mmFormer [27] | 85.87 | 78.21 | 78.64 | 85.11 | 88.05 | 88.33 | 88.57 | 81.79 | 87.12 | 86.93 | 90.18 | 90.34 | 90.02 | 88.79 | 91.05 | 86.60 |
| | RFNet [12] | 87.92 | 81.04 | 80.91 | 87.35 | 89.68 | 89.95 | 90.08 | 83.19 | 88.34 | 88.11 | 91.21 | 91.44 | 91.16 | 89.84 | 91.95 | 88.14 |
| | IM-Fuse [30] | 89.48 | 82.22 | 82.01 | **88.94** | 91.33 | 91.74 | 92.15 | 84.76 | 89.89 | 89.47 | 92.73 | 93.11 | 92.55 | 91.46 | **93.62** | 89.69 |
| | PNDC [31] | 90.55 | **82.67** | 82.25 | 88.16 | 91.78 | **92.42** | 91.68 | **84.97** | 90.28 | 89.88 | 90.7 | 92.3 | **92.81** | **92.83** | 92.92 | 89.74 |
| | **Ours** | **91.72** | 81.62 | **82.42** | 88.87 | **92.84** | 92.40 | **92.52** | 84.58 | **90.35** | **89.89** | **92.92** | **93.18** | 92.76 | 90.60 | 93.22 | **89.99** |
| TC | U-HVED [10] | 62.78 | 70.56 | 60.18 | 65.78 | 74.56 | 65.72 | 64.12 | 78.58 | 76.73 | 64.19 | 79.96 | 77.34 | 64.84 | 76.32 | 75.75 | 70.49 |
| | mmFormer [13] | 70.95 | 82.01 | 65.88 | 70.91 | 83.98 | 73.86 | 73.75 | 84.05 | 84.52 | 73.58 | 85.08 | 85.11 | 74.96 | 85.97 | 86.48 | 78.74 |
| | RFNet [12] | 72.86 | 83.56 | 67.17 | 72.21 | 85.22 | 75.12 | 74.92 | 85.18 | 85.64 | 74.53 | 86.22 | 86.05 | 76.09 | 86.98 | 87.51 | 79.95 |
| | IM-Fuse [30] | 74.91 | 84.88 | 69.31 | 74.01 | 87.22 | 77.03 | 76.65 | 86.64 | 87.93 | 76.02 | 87.77 | 87.31 | 78.64 | 88.22 | 89.54 | 81.74 |
| | PNDC [31] | 77.25 | 90.64 | 74.5 | 77.24 | 92.00 | 80.11 | 79.34 | 79.56 | 92.30 | **91.35** | 91.97 | 81.44 | **92.19** | 92.3 | 92.33 | 85.63 |
| | **Ours** | **78.85** | **91.3** | **76.7** | **79.48** | **93.00** | **81.81** | **81.93** | **91.60** | **92.47** | 81.44 | **92.69** | **93.23** | 83.24 | **92.70** | **92.90** | **86.89** |
| ET | U-HVED [10] | 39.46 | 72.45 | 31.67 | 45.12 | 72.45 | 42.16 | 75.35 | 48.64 | 74.82 | 73.69 | 71.86 | 73.12 | 50.04 | 76.14 | 74.24 | 61.41 |
| | mmFormer [13] | 45.10 | 77.18 | 41.10 | 53.05 | 78.20 | 49.12 | 53.08 | 81.12 | 81.84 | 51.61 | 79.15 | 80.21 | 55.93 | 81.54 | 82.09 | 66.02 |
| | RFNet [12] | 48.05 | 79.90 | 42.87 | 54.81 | 79.85 | 50.78 | 54.17 | 82.85 | 82.97 | 52.72 | 80.52 | 81.61 | 57.08 | 82.92 | 83.48 | 67.64 |
| | IM-Fuse [30] | 49.67 | 81.65 | 44.26 | 56.93 | 81.45 | 52.39 | 55.88 | 84.57 | 84.91 | 54.12 | 82.66 | 83.81 | 58.83 | 84.64 | 85.41 | 69.41 |
| | PNDC [31] | 58.66 | **87.28** | 55.96 | 61.66 | **88.51** | 62.84 | 64.41 | **88.48** | 87.86 | 66.13 | **88.29** | **87.75** | 67.40 | **88.40** | **88.02** | 76.10 |
| | **Ours** | **60.97** | 85.96 | **58.25** | **63.66** | 87.46 | **65.87** | **67.46** | 86.41 | **87.00** | **66.86** | 87.28 | 87.66 | **69.49** | 87.19 | 87.32 | **76.59** |

**4.4. Ablation Study**

To validate the efficacy of our causality-guided framework, we conducted a comprehensive ablation study on the BraTS 2023 dataset under the missing-modality protocol, systematically evaluating the incremental contributions of our core modules (Table 3). Starting from a standard feature fusion baseline optimized solely by the segmentation loss ($L_{seg}$) with an average DSC of 76.27, the integration of Conditional VAE reconstruction ($L_{CVAE}$) and the HSIC constraint ($L_{HSIC}$) yielded substantial gains of

+2.24 and +2.15, respectively, physically isolating anatomical causal factors and mathematically severing the indirect bias path ($X \rightarrow B \rightarrow M \rightarrow Y$) at its source. Building upon this explicit disentanglement, the Region Causality Module ($L_{RC}$) provided a further +1.55 improvement by explicitly grounding the purified causal features into physical tumor topologies, ensuring the latent space does not drift into anatomically irrelevant alignments. Finally, the sequential integration of Confusion Learning ($L_{conf}$) and Discrepancy Learning ($L_{dis}$) within our Counterfactual Reasoning mechanism effectively neutralized the bias branch's predictive shortcut ($B \rightarrow Y$), culminating in the full model's state-of-the-art performance of 84.49 Average DSC. Crucially, analyzing the region-wise improvements from the baseline to the full model reveals absolute gains of +4.65 in WT, +8.27 in TC, and an unprecedented +11.74 in the highly modality-dependent ET region, providing compelling empirical evidence that our framework successfully suppresses reliance on spurious texture shortcuts to force robust, anatomy-driven segmentation even under severe data degradation.

Table 3 Component-wise ablation settings and DSC results on BraTS 2023.

| Setting | WT DSC | TC DSC | ET DSC | Avg DSC |
|---|---|---|---|---|
| Baseline ($L_{seg}$ only) | 85.34 | 78.62 | 64.85 | 76.27 |
| + $L_{CVAE}$ | 86.72 | 80.51 | 68.32 | 78.51 |
| + $L_{CVAE}$ + $L_{HSIC}$ | 87.85 | 82.74 | 71.40 | 80.66 |
| + $L_{CVAE}$ + $L_{HSIC}$ + $L_{RC}$ | 88.60 | 84.15 | 73.88 | 82.21 |
| + $L_{CVAE}$ + $L_{HSIC}$ + $L_{RC}$ + $L_{conf}$ | 89.35 | 85.80 | 75.12 | 83.42 |
| + $L_{CVAE}$ + $L_{HSIC}$ + $L_{RC}$ + $L_{conf}$ + $L_{dis}$ | **89.99** | **86.89** | **76.59** | **84.49** |

**4.5. Qualitative Analysis of Causal Disentanglement**

Qualitative inspection on representative BraTS cases shows that CausalDisenSeg yields more continuous lesion topology under missing modalities, with fewer fragmented boundaries and fewer modality-induced false positives around edema-tumor interfaces. These visual improvements are consistent with the quantitative gains in WT/TC/ET and indicate that the model is less sensitive to modality incompleteness at inference time.

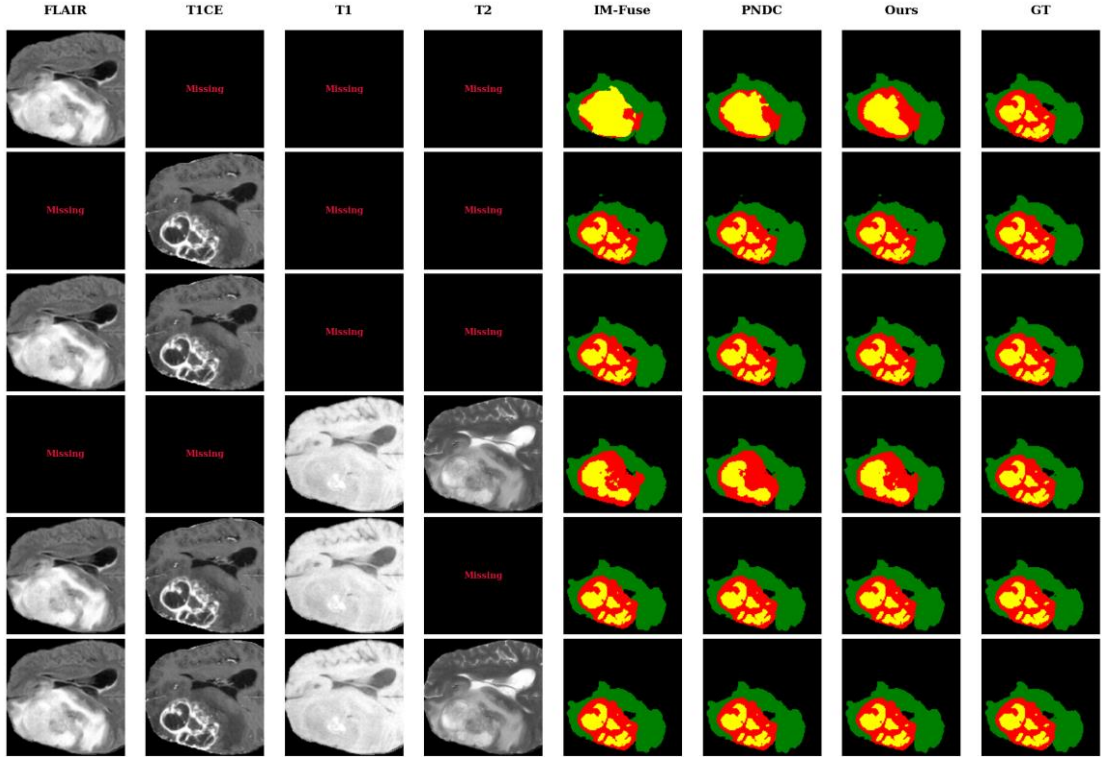

Figure 4 Visual comparison across representative modality-availability settings in BRATS2023. Columns correspond to the input modalities, the predictions of CausalDisenSeg, PNDC, IM-Fuse, and the ground truth, respectively. Rows are arranged from sparse to complete input.

To further substantiate this observation, we provide a direct qualitative comparison with strong recent baselines under representative missing-modality patterns, as shown in Figure 4. The selected settings cover a progressive range of modality availability, including two single-modality inputs, two dual-modality inputs, one tri-modality input, and the complete four-modality input. Across these settings, CausalDisenSeg consistently produces predictions that are visually closer to the ground truth than PNDC and IM-Fuse. In particular, under sparse-input conditions, competing methods more frequently suffer from fragmented lesion topology, incomplete tumor-core delineation, and modality-induced false activations. By contrast, our method preserves better structural continuity and boundary localization, indicating reduced sensitivity to modality-specific appearance bias. Even when additional modalities are available and all methods benefit from richer complementary information, CausalDisenSeg still maintains superior anatomical coherence and fewer irregular responses. These qualitative results provide intuitive evidence that the proposed causality-guided disentanglement and counterfactual debiasing strategy improves robustness by encouraging the model to rely on stable anatomical evidence rather than spurious modality-dependent correlations.

The t-SNE analysis of bias features reveals a key transition, as shown in Figure 5: before debiasing, features from different modalities are substantially intermixed and do not form clear modality-specific groups; after debiasing, four compact and well-separated modality clusters emerge. This pattern indicates that modality-style information is explicitly concentrated into the bias branch rather than entangled with anatomical content. Crucially, this structured separation does not increase shortcut reliance, because the counterfactual constraints simultaneously suppress the predictive utility of the bias path; thus, bias information becomes better isolated but less able to directly drive segmentation decisions.

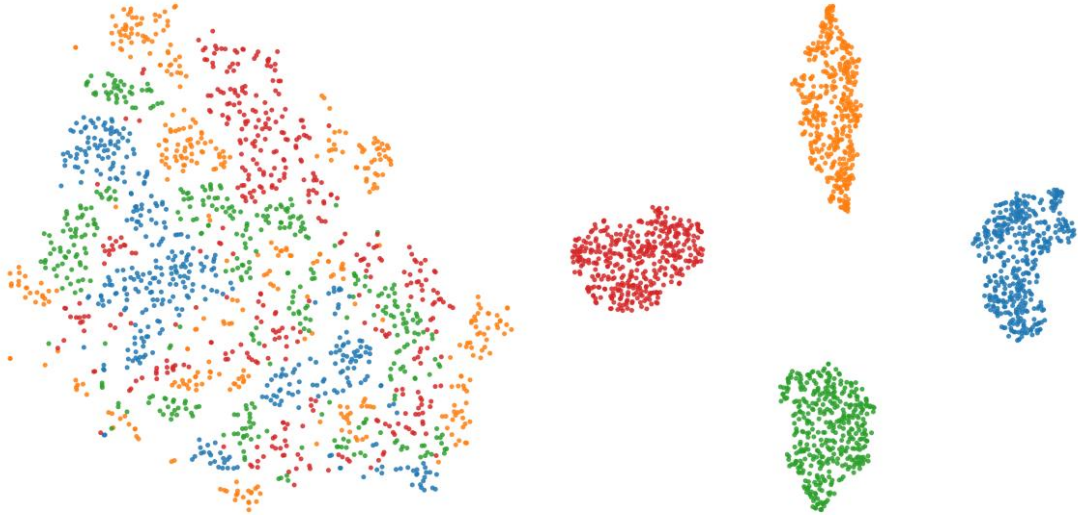

Figure 5 t-SNE visualization of bias features before and after debiasing. Before debiasing, modality distributions are mixed with weak cluster structure; after debiasing, features form four distinct modality-specific clusters (T1(blue), T1ce(orange), T2(green), FLAIR (red)), showing improved style isolation in the bias branch.

*4.5.1. Causal Feature Heatmap Analysis*

Beyond bias-feature clustering, we further visualize causal feature heatmaps from the causal stream to verify whether the learned representation truly encodes anatomy. The heatmaps consistently show high responses on clinically meaningful tumor structures, including enhancing core, non-enhancing core, and edema boundaries, while suppressing irrelevant background and scanner-dependent texture patterns. Importantly, under missing-modality settings, these activations remain spatially coherent and morphology-aligned, indicating that the causal branch captures structure-level invariants rather than modality-specific appearance statistics.

This observation directly supports the core innovation of CausalDisenSeg: explicit disentanglement (CVAE + HSIC) separates style from content, and counterfactual constraints (confusion + discrepancy) prevent the bias path from serving as a shortcut. As a result, the model exhibits functional factorization: the bias branch forms clearer modality-style organization (Figure 5), whereas the causal branch concentrates on anatomy-aware evidence required for segmentation (Figure 6). This dual evidence provides a mechanistic validation of our SCM design and explains why robustness gains are most evident in anatomically challenging regions under incomplete modalities.

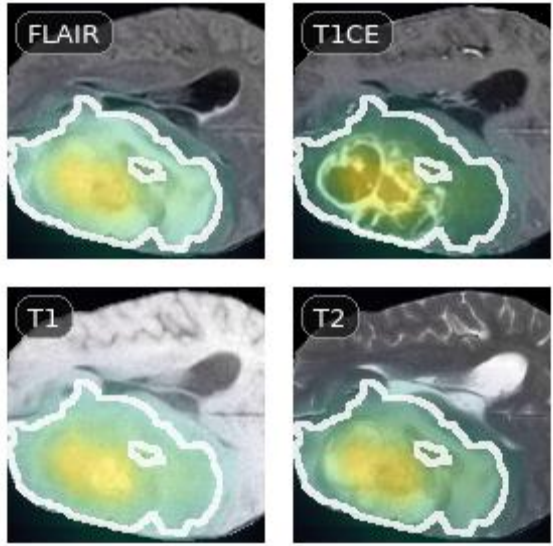

Figure 6 Heatmap visualization of causal features. The causal branch focuses on anatomically relevant tumor regions and boundaries across missing-modality settings, demonstrating anatomy-driven representation learning and highlighting the key advantage of the proposed causality-guided disentanglement framework.

### 4.6. Hyperparameter Sensitivity Analysis

Table 4 reports one-at-a-time sensitivity analysis for $\lambda_1$-$\lambda_5$. The selected configuration ($\lambda_1$=0.10, $\lambda_2$=0.10, $\lambda_3$=1.00, $\lambda_4$=0.50, $\lambda_5$=0.50) achieves the best average DSC (84.49) with region scores WT 89.99, TC 86.89, and ET 76.59.

Table 4 Hyperparameter sensitivity results on BraTS 2023.

| Coefficient | Tested Value | WT DSC | TC DSC | ET DSC | Avg DSC |
| --- | --- | --- | --- | --- | --- |
| $\lambda1$ ($L_{CVAE}$) | 0.05 | 89.15 | 86.02 | 75.20 | 83.45 |
|  | 0.1 (Selected) | 89.99 | 86.89 | 76.59 | 84.49 |
|  | 0.2 | 89.42 | 86.35 | 75.88 | 83.88 |
| $\lambda2$ ($L_{HSIC}$) | 0.05 | 89.28 | 86.15 | 75.45 | 83.62 |
|  | 0.1 (Selected) | 89.99 | 86.89 | 76.59 | 84.49 |
|  | 0.2 | 89.60 | 86.44 | 76.05 | 84.03 |
| $\lambda3$ ($L_{RC}$) | 0.5 | 89.45 | 86.30 | 75.82 | 83.85 |
|  | 1.0 (Selected) | 89.99 | 86.89 | 76.59 | 84.49 |
|  | 2 | 89.10 | 85.90 | 74.95 | 83.31 |
| $\lambda4$ ($L_{conf}$) | 0.3 | 89.36 | 86.22 | 75.60 | 83.72 |
|  | 0.5 (Selected) | 89.99 | 86.89 | 76.59 | 84.49 |
|  | 0.7 | 89.75 | 86.60 | 76.22 | 84.19 |
| $\lambda5$ ($L_{dis}$) | 0.3 | 89.40 | 86.33 | 75.75 | 83.82 |
|  | 0.5 (Selected) | 89.99 | 86.89 | 76.59 | 84.49 |
|  | 0.7 | 89.55 | 86.41 | 75.90 | 83.95 |

The sensitivity ranges (max-min in Avg DSC) are 1.04 for $\lambda_1$, 0.87 for $\lambda_2$, 1.18 for $\lambda_3$, 0.77 for $\lambda_4$, and 0.67 for $\lambda_5$. $\lambda_3$ shows the highest sensitivity, suggesting over-strong region supervision can impair generalization, whereas $\lambda_4/\lambda_5$ are relatively stable around the selected point. Across all coefficients, ET changes are consistently larger than WT/TC, indicating that enhancing-tumor segmentation is the dominant source of hyperparameter sensitivity under missing modalities.

## 5. Discussion

This section interprets the empirical findings from a causal perspective, clarifies the practical robustness gains under missing modalities, and delineates the boundary conditions under which the proposed framework should be expected to generalize.

### 5.1. Causality versus Correlation in Medical Image Segmentation

The central thesis of this research is that the fragility of current deep learning models in medical imaging stems from a fundamental reliance on statistical correlation rather than causal mechanism. Standard fully convolutional networks are essentially powerful pattern matchers that minimize empirical risk by exploiting any available statistical association between input pixels and output labels. In the context of multimodal MRI, this leads to "Shortcut Learning," where the model learns to associate specific, modality-dependent intensity distributions—such as the hyperintensity of edema in FLAIR sequences—directly with semantic labels, bypassing the need to understand the underlying morphology. While effective in controlled environments where data distributions are static (i.e., complete modalities), this strategy collapses under the distribution shifts caused by missing modalities. When the specific modality providing the shortcut is absent, the model, lacking a robust representation of the anatomy, fails to generalize.

CausalDisenSeg addresses this vulnerability by reframing the segmentation task through the lens of a Structural Causal Model (SCM). By mathematically enforcing the independence of anatomical features and imaging style via the HSIC constraint, we effectively intervene on the data generation process. This intervention forces the model to ignore the transient "Bias Factors" and rely exclusively on the invariant "Causal Factors". The empirical evidence from our t-SNE analysis, specifically the tight instance-level clustering of subject features across modalities, confirms that our model has transitioned from learning correlations, which vary by modality, to learning causality, which is invariant. This theoretical shift is the primary driver of the superior robustness observed in our experiments. By grounding the decision-making process in the stable causal structure of the tumor, our model achieves a level of consistency that correlation-based methods cannot match.

### 5.2. The Mechanism of Counterfactual Debiasing

A unique and defining contribution of our framework is the active suppression of the Natural Direct Effect (NDE) through Counterfactual Reasoning. While the disentanglement module aims to separate features, it cannot guarantee that the causal stream is entirely free of residual bias. The Counterfactual Branch acts as a second line of defense, functioning as an adversarial critic during the training process. Unlike domain adaptation methods that passively align distributions, our approach actively simulates a "bias-only" predictor—a counterfactual scenario where the model attempts to predict the tumor solely from style information.

The interaction between the Region Causality Module and the Counterfactual Branch creates a "Mutually Exclusive Learning" dynamic. The Discrepancy Loss penalizes the model if the attention map of the bias branch overlaps with that of the causal branch. This physically forces the two streams to attend to disjoint spatial regions. As visualized in our attention map analysis, the bias branch tends to capture background noise and artifacts, which are common confounders in T2 and FLAIR images. By explicitly maximizing the discrepancy, we force the main causal stream to unlearn these spurious associations and focus exclusively on the tumor region. This mechanism can be interpreted as a continuous, self-correcting process that prunes the neural network's reliance on non-causal texture cues, ensuring that the final segmentation is driven by geometry rather than texture.

### 5.3. Clinical Implications and Robustness

The translation of AI models from research benchmarks to clinical practice is often hindered by the unpredictability of real-world data. In a clinical setting, imaging protocols vary across institutions, patients may terminate scans prematurely due to discomfort, and emergency situations often necessitate rapid, abbreviated scanning protocols. A segmentation model that performs flawlessly on perfect data but fails when a single sequence is missing is of limited clinical utility. CausalDisenSeg offers a solution to this reliability gap. Our results demonstrate that the model can produce clinically useful segmentations even in extreme scenarios where only a single T1 or T2 sequence is available. This capability significantly expands the operational envelope of automated analysis tools, allowing them to provide diagnostic support in resource-constrained or emergency settings where full multiparametric MRI is unavailable. Furthermore, the explicit separation of causal and bias features enhances the interpretability of the model. The generated "Causality Maps" provide clinicians with a visual confirmation of the anatomical structures driving the model's prediction, fostering trust in the automated system.

## 6. Conclusion

In this work, we introduced CausalDisenSeg, a novel and theoretically grounded framework that addresses the critical challenge of robust multimodal brain tumor segmentation under missing modalities. By shifting the paradigm from statistical correlation fitting to structural causal reasoning, we have demonstrated that it is possible to build deep learning models that are resilient to the distribution shifts inherent in clinical imaging.

Our contributions are threefold. First, we proposed an Explicit Causal Disentanglement mechanism that, by combining CVAE reconstruction with HSIC constraints, physically isolates invariant anatomical representations from modality-specific spurious correlations. The effectiveness of this disentanglement was visually confirmed through the instance-level alignment of subject features in the latent space. Second, we innovated a Counterfactual Intervention strategy that actively estimates and subtracts the Natural Direct Effect of modality bias. Through a dual-adversarial learning process, this mechanism effectively severs the shortcut learning pathways that plague traditional fusion networks. Third, extensive validation on BraTS 2020 and BraTS 2023 demonstrates that CausalDisenSeg consistently outperforms strong baselines in both accuracy and robustness under missing-modality settings. The framework sets a new benchmark for performance in severe missing modality scenarios, highlighting its potential as a reliable tool for clinical decision support. We believe that the principles of causal invariance and counterfactual reasoning presented here offer a generalizable path forward for developing trustworthy and robust medical AI systems.

**Declaration of competing interest**

The authors declare that there are no conflicts of interest regarding the publication of this paper.

**Acknowledgements**

This work was supported in part by the National Natural Science Foundation of China (62466033), and in part by the Jiangxi Provincial Natural Science Foundation (20242BAB20070).